\documentclass[letterpaper]{article} 
\usepackage{aaai2026}  
\usepackage{times}  
\usepackage{helvet}  
\usepackage{courier}  
\usepackage[hyphens]{url}  
\usepackage{graphicx} 
\usepackage{amsmath}
\usepackage{graphicx}
\graphicspath{{images/}}
\usepackage{booktabs}   
\usepackage{tabularx}   
\usepackage{array}      
\usepackage{ragged2e}   
\usepackage{subcaption}
\usepackage{xcolor}

\newcommand{\compactcell}[1]{\begin{minipage}[t]{\linewidth}\small\RaggedRight #1\end{minipage}}

\usepackage{natbib}  
\usepackage{caption} 
\usepackage{algorithm}
\usepackage{algorithmic}
\usepackage{multirow}
\usepackage{newfloat}
\usepackage{listings}
\DeclareCaptionStyle{ruled}{labelfont=normalfont,labelsep=colon,strut=off} 
\floatstyle{ruled}
\newfloat{listing}{tb}{lst}{}
\floatname{listing}{Listing}
\title{Interpreting Fedspeak with Confidence: A LLM-Based Uncertainty-Aware Framework Guided by Monetary Policy Transmission Paths}

\author {
    Rui Yao\equalcontrib\textsuperscript{\rm 1},
    Qi Chai\equalcontrib\textsuperscript{\rm 1},
    Jinhai Yao\equalcontrib\textsuperscript{\rm 2},
    Siyuan Li\textsuperscript{\rm 1},
    Junhao Chen\textsuperscript{\rm 1},
    Qi Zhang\thanks{Corresponding author.}\textsuperscript{\rm 2},
    Hao Wang\footnotemark[\value{footnote}]\textsuperscript{\rm 1}
}
\affiliations {
    \textsuperscript{\rm 1}The Hong Kong University of Science and Technology (Guangzhou)\\
    \textsuperscript{\rm 2}Antai College of Economics and Management, Shanghai Jiaotong University\\
    \{ryao663, qchai315, sli974, jchen024\}@connect.hkust-gz.edu.cn,
    \{jh\_Yao,zhang.qi\}@sjtu.edu.cn,
    haowang@hkust-gz.edu.cn
}

\begin{document}
\nocopyright
\maketitle

\begin{abstract}
``Fedspeak", the stylized and often nuanced language used by the U.S. Federal Reserve, encodes implicit policy signals and strategic stances. The Federal Open Market Committee  strategically employs Fedspeak as a communication tool to shape market expectations and influence both domestic and global economic conditions. As such, automatically parsing and interpreting Fedspeak presents a high-impact challenge, with significant implications for financial forecasting, algorithmic trading, and data-driven policy analysis.
In this paper, we propose an LLM-based, uncertainty-aware framework for deciphering Fedspeak and classifying its underlying monetary policy stance.
Technically, to enrich the semantic and contextual representation of Fedspeak texts, we incorporate domain-specific reasoning grounded in the monetary policy transmission mechanism.
We further introduce a dynamic uncertainty decoding module to assess the confidence of model predictions, thereby enhancing both classification accuracy and model reliability.
Experimental results demonstrate that our framework achieves state-of-the-art performance on the policy stance analysis task.
Moreover, statistical analysis reveals a significant positive correlation between perceptual uncertainty and model error rates, validating the effectiveness of perceptual uncertainty as a diagnostic signal.
\end{abstract}


\begin{links}
    \link{Code}{https://github.com/yuuki20001/FOMC-sentiment-path}
\end{links}

\section{Introduction}

The Federal Open Market Committee (FOMC) is the key institution responsible for managing the U.S. economy and implementing monetary policy~\cite{CONNOLLY2024107286}. Through tools such as interest rate adjustments and open market operations, monetary policy regulates the money supply to influence economic activity~\cite{PFLUEGER202271, VOLK2024104}. FOMC’s statutory mandate is to promote ``maximum employment, stable prices, and moderate long-term interest rates". This reflects a broader goal of maintaining a sustainable balance between economic growth, labor market performance, and inflation control. Given the global importance of the Federal Reserve, its communications have a direct and substantial influence on financial markets worldwide~\cite{ KARNAUKH202255, SUH2025107040}.

Federal Reserve communications, known as ``Fedspeak", are challenging to interpret. This challenge arises from their inherent ambiguity, where the same term may imply different policy stances depending on the broader economic context. For example, a ``strong" labor market might be a dovish signal (indicating no immediate rate hikes) in a weak economy, but it can become a hawkish signal (suggesting tightening) in an overheated economy. This strong dependence on context poses a major challenge for traditional sentiment analysis models.

Traditional methods in financial sentiment analysis often face a trade-off between performance and interpretability. For instance, dictionary-based approaches are simple and highly interpretable, but their performance is often limited as they struggle to understand complex contexts or incorporate external knowledge. In contrast, language models like FinBERT~\cite{araci2019finbert,liu2021finbert}, after being fine-tuned with domain-specific supervised data~\cite{shah2023trillion}, can capture contextual nuances effectively and achieve superior performance. However, the black-box nature of such model leads to a lack of transparency in their decision-making process, resulting in poor interpretability. 

\begin{figure}[t]
    \centering
    \begin{subfigure}[b]{0.8\columnwidth}
        \centering
        \includegraphics[width=\linewidth]{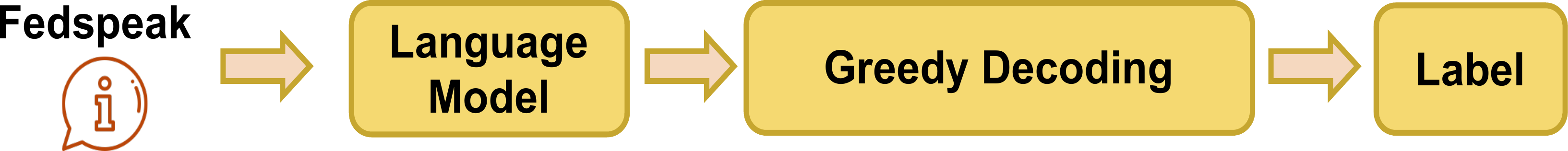}
        \caption{Traditional Method}
        \label{fig:traditional}
    \end{subfigure}
    
    
    \begin{subfigure}[b]{0.8\columnwidth}
        \centering
        \includegraphics[width=\linewidth]{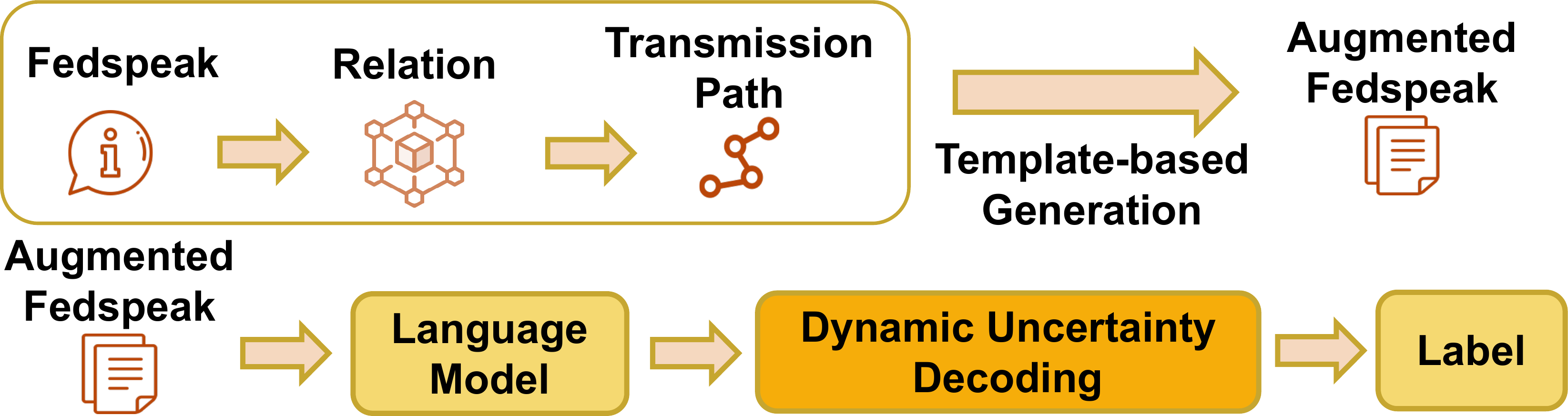}
        \caption{Our Method}
        \label{fig:our}
    \end{subfigure}

    \caption{\textbf{Comparison between (a) the traditional method and (b) our proposed method}. Our method introduces a domain-specific reasoning approach grounded in the monetary policy transmission mechanism to emulate how human experts analyze policy stances, providing the model with relevant domain knowledge. In addition, we employ a dynamic uncertainty decoding module to capture perceptual uncertainty and estimate the model's confidence in its predictions, thereby improving overall reliability.}

    \label{fig:comparison}
\end{figure}

LLMs offer an accessible and effective solution for financial sentiment analysis, particularly for deciphering central bank policy stances. Recent studies~\cite{peskoff2023gpt, hansen2024can} have shown that closed-source LLMs, such as the GPT-4 series, exhibit strong zero-shot capabilities on policy stance analysis tasks. Research from central banks~\cite{gambacorta2024cb,geiger2025monetary} demonstrates that fine-tuned LLMs can match human-level accuracy in interpreting monetary policy stance signals. However, existing work primarily focuses on performance metrics, often neglecting critical aspects of LLM behavior such as reliability, bias, and hallucinations. Improving the reliability and interpretability of LLMs, as well as understanding their limitations in policy stance analysis, is essential for reducing systemic risks and enhancing financial stability~\cite{leitner2024rise}.

In economics, uncertainty is often categorized as risk (known probabilities) or ambiguity (unknown probabilities). Recent work in behavioral economics borrows the computer science concepts of epistemic and aleatoric uncertainty to describe how investors interpret uncertainty~\cite{walters2023investor}. Epistemic uncertainty reflects missing knowledge, while aleatoric uncertainty stems from randomness. This framing aligns closely with how LLMs operate: like investors, they make predictions under incomplete information. We adopt this perspective to quantify model uncertainty and use it to identify the confidence of predictions. We build on this analogy and treat LLMs as policy stance analysts. 

Therefore, to align the model behavior with that of human analysts, we augment the input texts by extracting financial entity relations and reasoning over monetary policy transmission paths using structured templates. This augmentation is designed both to emulate expert-level economic reasoning and to reduce the model's cognitive risk by compensating for missing domain-specific knowledge.
Then, we redefine the model prediction uncertainty as perceptual uncertainty (PU), decomposed into cognitive risk (CR) and environmental ambiguity (EA). Building on this formulation, we introduce a dynamic uncertainty decoding module that adapts model behavior based on the PU level. We tune PU–related hyperparameters on the validation set and use them to evaluate the model prediction confidence.

\begin{figure*}[t]
\centering
\includegraphics[width=0.95\textwidth]{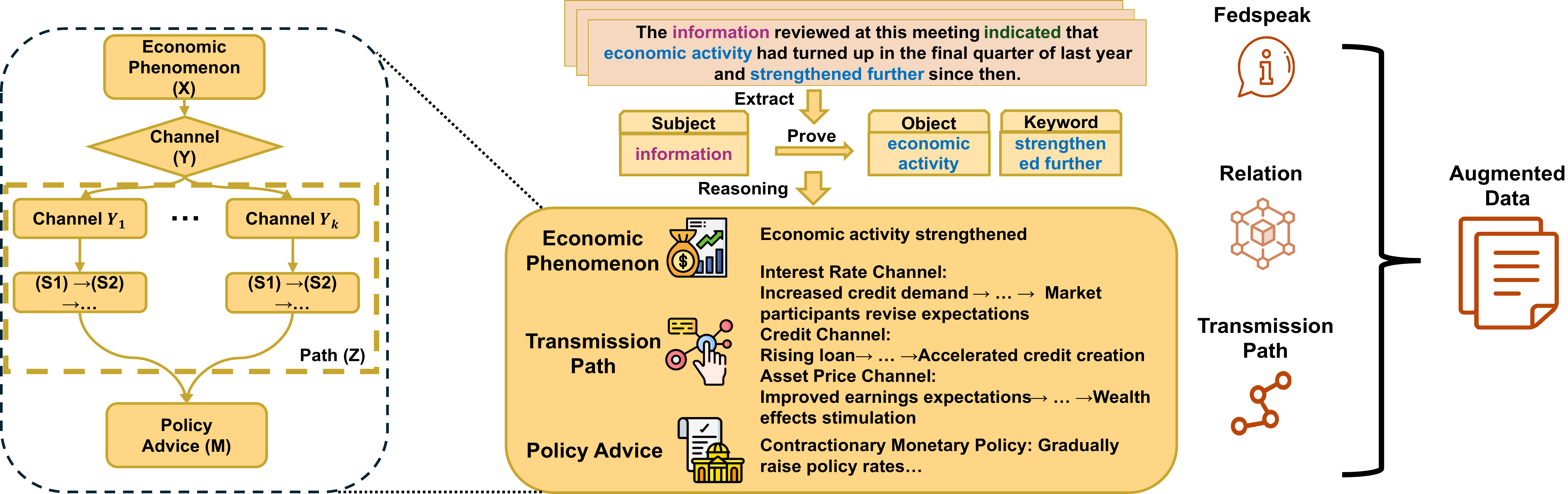}
\caption{\textbf{The workflow of data augmentation}.We extract economic entity relations from Fedspeak, and then perform reasoning grounded in the monetary policy transmission mechanism using structured templates to derive policy advice.}
\label{fig:data_augmentation}
\end{figure*}

The contribution of this work is as follows:
\begin{itemize}

\item We incorporate a domain-specific reasoning grounded in the monetary policy transmission mechanism to emulate the analytical process of human experts and enhance economic interpretability on the policy stance analysis task.
\item We introduce a dynamic uncertainty decoding module with a PU metric that helps identify potentially unreliable predictions, aiming to improve overall prediction reliability.
\item Our framework achieves state-of-the-art performance while enhancing transparency, economic interpretability, and human-AI collaboration in policy stance analysis.

\end{itemize}


\section{Related Work}

\subsection{AI in Finance}
Recent applications of LLMs in the financial domain have led to the emergence of domain-specific models. proprietary models like BloombergGPT~\cite{wu2023bloomberggpt} and open-source models such as FinGPT~\cite{liu2023fingpt} are often instruction-tuned to perform financial tasks such as sentiment analysis. Existing financial evaluation benchmarks, such as FinQA~\cite{chen2021finqa}, FinBen~\cite{xie2024finben}, FinTextQA~\cite{chen-etal-2024-fintextqa}, and FinDER~\cite{choi2025finder}, are designed to assess the performance of LLMs on a diverse range of tasks. These tasks cover numerical reasoning, retrieval-augmented generation evaluation, general understanding, and financial regulation and compliance. These benchmarks establish a standard for measuring the reliability of LLMs in various financial tasks.
Another line of research enhances LLM reasoning by decomposing complex and ambiguous tasks into structured, domain-informed components, as seen in FinEntity, EFSA, and DEFINE~\cite{tang2023finentity,chen2024efsa,hu2024define}. These works highlight the value of integrating domain priors with structured task design to improve accuracy and reliability.

\subsection{FOMC Analysis}
Fedspeak plays a pivotal role in shaping market expectations and asset prices, offering key insights into the Federal Reserve's economic outlook and policy intentions~\cite{ehrmann2020starting,cieslak2021economics, mathur2022monopoly,gomez2022real}. Its nuanced language requires careful interpretation, as tone and sentiment in communications can significantly affect financial variables and market volatility~\cite{gorodnichenko2023voice,curti2023let}. Approaches to analyzing Fedspeak have evolved from early dictionary-based methods~\cite{loughran2011liability} and topic modeling techniques~\cite{boukus2006information,edison2021text} to Transformer-based models that better capture contextual subtleties~\cite{liu2023fingpt,shah2023trillion}. LLM-based methods have become increasingly popular for policy stance analysis~\cite{peskoff2023gpt,hansen2024can}. Domain-specific adaptations have shown promise in improving classification accuracy~\cite{gambacorta2024cb,geiger2025monetary}. However, most existing approaches limit LLMs to predicting stance labels, underutilizing their potential.


\section{Methodology}
\subsection{Policy Stance Definition}
When setting monetary policy, the Federal Reserve relies on deliberations and votes by FOMC members to reach collective decisions. These decisions reflect an aggregation of individual views and preferences. Committee members hold different policy stances: Members who prefer supporting increased output and employment and exhibit a stronger aversion to unemployment risk are considered dovish. Members who prefer controlling inflation and exhibit stronger aversion to inflation risk are considered hawkish~\cite{BORDO2023125}. From a trade-off perspective, dovish members place more weight on maximizing employment, whereas hawkish members assign more weight to maintaining price stability~\cite{shah2023trillion, SMALES2023104514}. We follow prior literature in assigning stance labels based on these definitions. We define policy stances as dovish, hawkish, and neutral. The detailed definitions and  explanations are illustrated in the detailed data augmentation section of the appendix.

\subsection{Data Augmentation with Domain Reasoning}
\subsubsection{Financial Entity Relations}
The relations embedded in Fedspeak are complex and multi-layered. To enhance the reliability of interpreting Fedspeak, we decompose these financial entity relations, thereby improving our model’s reasoning capability and accuracy~\cite{10.1145/3677052.3698603}. Within the financial entity relations analysis framework, atomic relations serve as the minimal logical units for semantic decomposition. We define the entity set as $\mathcal{E}$ (representing all financial and economic entities) and the atomic relation set as $\mathcal{R} = \{CAUSE, COND, EVID, PURP, ACT, COMP\}$. Atomic relations are formally defined in Equation \ref{eq: Atomic Relation}:
\begin{equation}
r(e_i,e_j) \in  \mathcal{R} \quad \text{,} \quad \forall e_i,e_j \in \mathcal{E}
\label{eq: Atomic Relation}
\end{equation}
where $e_i$ is the subject entity, $e_j$ is the object entity, and $r$ is one of the six core relation types. These six core relation types are listed in the detailed data augmentation section of the appendix.\par

\begin{figure*}[htp]
\centering
\includegraphics[width=1\textwidth]{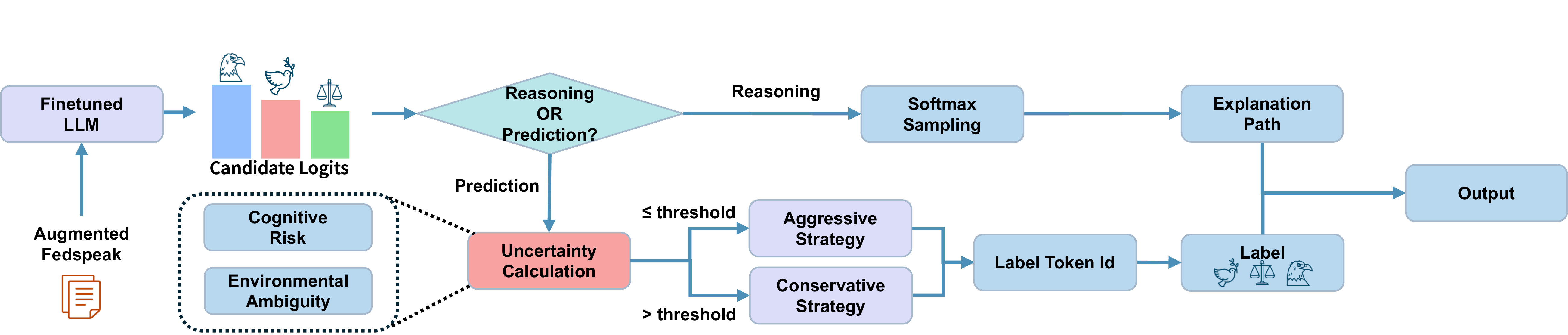}
\caption{\textbf{Overview of Dynamic Uncertainty Decoding module}.When the LLM is about to generate a prediction token, we obtain the corresponding logits over the vocabulary. Dynamic uncertainty decoding module quantifies the model's PU via estimating CR and EA. The decoding strategy for the current token is selected based on whether the PU exceeds the threshold. }
\label{fig:Dynamic Uncertainty Decoding}
\end{figure*}

Given a path $T:e_1 \xrightarrow{r_1} e_2 \xrightarrow{r_2} \dots \xrightarrow{r_n} e_{n+1}$, the decomposition of financial entity relations follows Equation \ref{eq: the decomposition of financial entity relations}:
\begin{equation}
T \Rightarrow \bigcup_{k=1}^{n} r_k(e_k, e_{k+1}) \quad \text{,} \quad r_k \in \mathcal{R}
\label{eq: the decomposition of financial entity relations}
\end{equation}
In Equation \ref{eq: the decomposition of financial entity relations}, if a sentence contains a multi-step logical chain or causal chain (e.g., A causes B, B causes C), it must be decomposed into a sequential list of atomic relations within the same subtask. We construct an entity classification function $C:\mathcal{E}\rightarrow E$ that maps to the information sources $E$ in Fedspeak. When extracting entity relations, the information source (e.g., official sources: Fed chairs, official documents, committee members; external sources: journalists, analysts) and its stance (official statements, data interpretation, external analysis, direct questions, rhetorical questions) must be identified.


\subsubsection{Monetary Policy Transmission Paths}
Existing research on monetary policy primarily focuses on understanding how policy tools, such as interest rate adjustments, affect key economic targets like inflation and employment. These studies trace the monetary policy transmission paths through the financial system and the real economy, offering a theoretical foundation for policy design~\cite{PASSOS2024101923, BOSSHARDT2024103914, SUH2025107040,  ALESSANDRI2025103783}.
 
Given the importance of monetary policy transmission paths, we construct the transmission path: Define a quadruple $\mathbf{\Gamma} = (\mathbf{X}, \mathbf{Y}, \mathbf{Z}, \mathbf{M})$, where $\mathbf{X} \in{R}^n$ is an $n$ dimensional shock vector representing economic phenomena, policy implementation, policy framework shifts, or external shocks, serving as the starting point; $\mathbf{Y} = \{y_1, y_2, \dots, y_k\}$ is the set of monetary policy transmission channels, with $k$ being the number of channels, including the credit channel, asset price channel, aggregate demand channel, etc.; $\mathbf{Z} = \{Z_1, Z_2, \dots, Z_p\}$ is the monetary policy transmission path, defined as the sequence of expectation shifts or market/economic indicator responses within a specific transmission channel. For each channel $y_i \in \mathbf{Y}$, there exists a transmission path $Z_{y_{\text{i}}}$, which consists of a series of state transitions, as shown in Equation \ref{eq: return1}:
\begin{equation}
Z_{y_{\text{i}}} = f_{\text{i,n}} \circ f_{\text{i,n-1}} \circ \cdots \circ f_{\text{i,j}}\circ \cdots \circ f_{\text{i,1}}
\label{eq: return1}
\end{equation}
where $f_{\text{i,j}}$ denotes the mapping function at step $j$ in channel $y_i$. For each channel $y_i$, the state sequence is given by Equation \ref{eq: return2}:
\begin{equation}
S_{\text{i},j} =
\begin{cases} 
\phi_i(\mathbf{X}) \quad j=i^{(0)}\\
g^{(j)}_{\text{i},j}(S_{\text{i},j}) \quad j=i^{(1)},i^{(2)},\dots,i^{(n)}
\end{cases}
\label{eq: return2}
\end{equation}
In Equation \ref{eq: return2}, $j$ denotes the number of steps in channel $y_i$, $\phi_i$ represents the function that maps the initial shock $\mathbf{X}$ to the initial state of this channel, and $S_{\text{i},n_i}$ is the final state of the channel. $\mathbf{M} \in \mathcal{\mathbf{P}}$ is the policy advice, $\mathbf{P}$ denotes the policy space, and the entire transmission path of channel $y_i$ is aggregated into $S^{(i)} = \{S_{\text{i},0}, S_{\text{i},1}, \dots, S_{\text{i},n_i}\}$. The policy advice $\mathbf{M}$ is a function of the aggregated state, as given by Equation \ref{eq: return3}:
\begin{equation}
\mathbf{M} = h\left(S^{(1)},S^{(2)}, \dots,S^{(i)}, \dots,S^{(k)} \right)
\label{eq: return3}
\end{equation}
where the final policy advice is generated by aggregating channel outputs and influenced by multiple channels and transmission paths. Through the above analysis, we adopt a structured framework to analyze the monetary policy transmission mechanism, with Equation \ref{eq: return4} as the main path. 
\begin{equation}
\mathbf{X} \xrightarrow\ \mathbf{Z}\xrightarrow\ \mathbf{M}
\label{eq: return4}
\end{equation}
The detailed transmission path of monetary policy is illustrated in Figure~\ref{fig:data_augmentation}.

\subsection{Dynamic Uncertainty Decoding}
LLMs may generate unreliable information. During next-token prediction, reliability can be affected by ambiguity in the input data and insufficient domain knowledge.
We treat our model as a policy analyst. Policy analysts face reliability issues in decision-making due to EA and CR. 
EA stems from data-related uncertainty, such as distributional shifts, semantic ambiguity, and other stochastic variations in the input~\cite{walters2023investor, LEROUX2025102323}.
EA causes policy analysts to consider multiple alternative options during decision-making.
CR arises from limited domain knowledge, missing information, or skill gaps~\cite{walters2023investor}.
CR causes policy analysts to make unreliable judgments due to cognitive limitations. Higher levels of EA or CR increase the policy analyst's PU. We incorporate a measure of PU into our model framework. 
Let the LLM be denoted as $\mathcal{M}$, and let the prompt be tokenized into vector representations $\mathbf{q}$. 
We use the top-$k$ logits from $\mathcal{M}$ to construct a Dirichlet distribution~\cite{sensoy2018evidential, ma2025estimating}, as shown below:
\begin{equation}
\alpha_k = \mathcal{M}(\tau_k|\mathbf{q},\mathbf{a_{t-1}}) \quad \text{,} \quad \alpha_0=\sum_{k=1}^K\alpha_k
\label{eq: Dirichlet distribution}
\end{equation}

where $\tau_k$ presents the token associated with the $k^{th}$ largest logit, and $\alpha_0$ constitutes the total evidence parameter of the Dirichlet distribution, equaling the sum of the $k$ largest logits. The model predicts the next token $a_t$ based on the representations $\mathbf{q}$ and the previously generated tokens
$\mathbf{a_{t-1}}=a_1a_2\dots a_{t-1}$. \par
Our measure of EA is defined as follows:
\begin{equation}
EA(a_t) = -\sum_{k=1}^K\frac{\alpha_k}{\alpha_0}(\psi(\alpha_k+1)-\psi(\alpha_0+1))
\label{eq: EA}
\end{equation}
where $\psi$ represents the digamma function, defined as $\psi(x)=\frac{d}{dx}\ln\Gamma(x)$. EA denotes the expected entropy of the predictive distribution. Higher entropy indicates a more uniform data distribution and greater ambiguity. Increased EA implies higher inherent uncertainty in the data, making the model uncertain about the input content. In such cases, a high level of EA indicates that the model is uncertain and confused about the output.\par

Our measure of CR is defined as follows:
\begin{equation}
CR(a_t) = \frac{K}{\sum_{k=1}^K(\alpha_k+1)}
\label{eq: CR}
\end{equation}
CR is inversely related to the total evidence $\sum_{k=1}^K(\alpha_k+1)$. A higher CR indicates that the model has less accumulated evidence and domain knowledge, suggesting limited capability to process the input effectively. Under high CR, the model tends to make less reliable decisions.\par

Our measure of PU is defined as follows:
\begin{equation}
PU =EA \times CR
\label{eq: PU}
\end{equation}
where PU is jointly determined by EA and CR. When both EA and CR values are high, the model struggles to identify clear information in the input and lacks sufficient domain knowledge for accurate judgment. This results in elevated PU and less reliable outputs. To enhance the model’s understanding of policy analysis and improve its reliability, we integrate uncertainty quantification into our framework with the goal of reducing PU.

\begin{table*}[ht]
\centering
\small
\setlength{\tabcolsep}{1.9mm}
\begin{tabular}{lcccccccc}
  \toprule
  \multirow{2.5}{*}{\textbf{Model}} & \multicolumn{2}{c}{\textbf{Meeting Minutes}} & \multicolumn{2}{c}{\textbf{Press Conference}} & \multicolumn{2}{c}{\textbf{Speeches}} & \multicolumn{2}{c}{\textbf{All Categories}} \\ 
  \cmidrule{2-9}
   & \textbf{Macro F1} & \textbf{Weighted F1} & \textbf{Macro F1} & \textbf{Weighted F1} & \textbf{Macro F1} & \textbf{Weighted F1} & \textbf{Macro F1} & \textbf{Weighted F1} \\ 
  \midrule
  \multicolumn{9}{l}{\textit{\textbf{Zero-Shot}}}\\

  ~GLM-4-9B & 0.2855 & 0.3058 & 0.3438 & 0.3876 & 0.3065 & 0.3807 & 0.3152 & 0.3478\\ 
  ~HD-Dissent & 0.3032 & 0.4219 & 0.3359 & 0.5962 & 0.2468 & 0.3288 & 0.3109 & 0.4774\\

  ~Qwen3-8B & 0.5144 & 0.5325 & 0.5670 & 0.5843 & 0.5748 & 0.6450 & 0.5572 & 0.5854\\ 
  ~Qwen3-14B & 0.5083 & 0.5367 & 0.4236 & 0.4650 & 0.5129 & 0.5969 & 0.5135 & 0.5514\\
  ~GLM-Z1-9B & 0.5998 & 0.5995 & 0.5951 & 0.6143 & 0.5761 & 0.6284 & 0.6023 & 0.6154\\ 
  ~Deepseek-R1 & 0.6269 & 0.6355 & 0.5873 & 0.6094 & 0.5680 & 0.6490 & 0.6201 & 0.6385\\
  ~Gemini-2.5-pro & 0.5953 & 0.5823 & 0.6703 & 0.6620 & 0.6116 & 0.6550 & 0.6286 & 0.6275 \\ 
  ~Phi-4 & 0.6366 & 0.6385 & 0.6023 & 0.6105 & 0.6140 & 0.6736 & 0.6349 & 0.6488\\ 
  ~GPT-4.1 & 0.6500 & 0.6456 & \textbf{0.6803} & \textbf{0.6900} & 0.6326 & 0.6980 & 0.6662 & 0.6763\\
  ~AICBC & 0.6701 & 0.6741& 0.6735 & 0.6805 & 0.6175 & 0.6841 & 0.6637 & 0.6802\\
  \midrule
  \multicolumn{9}{l}{\textit{\textbf{Fine-Tuned}}}\\
  ~GLM-4-9B & 0.6442 & 0.6514 & 0.5955 & 0.6087 & 0.6146 & 0.6832 & 0.6390 & 0.6585 \\ 
 
  ~Phi-4 & 0.6481 & 0.6560 & 0.5706 & 0.5927 & 0.6609 & 0.7096 & 0.6495 & 0.6683\\ 
  ~FinBERT & 0.5922 & 0.5958 & 0.6117 & 0.6844 & 0.6398 & 0.6513 & 0.6185 & 0.6380 \\ 
  ~GLM-Z1-9B & 0.6426 & 0.6514 & 0.6366 & 0.6503 & 0.6527 & 0.7032 & 0.6573 & 0.6736\\  
  ~Qwen3-8B & 0.6504 & 0.6536 & 0.5783 & 0.5985 & 0.6621 & 0.7242 & 0.6586 & 0.6745\\ 
  ~Qwen3-14B & 0.6227 & 0.6288 & 0.6286 & 0.6363 & 0.6052 & 0.6819 & 0.6360 & 0.6534\\

  \midrule
  \textbf{Ours} & \textbf{0.7449} & \textbf{0.7394} & 0.6672 &  0.6699 & \textbf{0.7291} & \textbf{0.7718} & \textbf{0.7327} & \textbf{0.7426} \\
    \bottomrule

\end{tabular}
\caption{Performance comparison of our method against zero-shot and fine-tuned baselines on the FOMC dataset. \textit{\textbf{Zero-Shot}} denotes the baseline models; \textit{\textbf{Fine-tuned}} applies LoRA fine-tuning on FOMC data to the \textit{\textbf{Zero-Shot}} models; \textbf{Ours} further integrates the proposed data augmentation and uncertainty decoding module with the \textit{\textbf{Fine-tuned}} Qwen-3-14B. }
\label{table:main}
\end{table*}

\section{Experiment}
\subsection{Experiment Setup}
To simulate the reasoning process of human analysts, we design a set of structured templates grounded in the mechanism of monetary policy transmission. Using these templates, we augment the original Fedspeak texts through a hybrid human-AI procedure to construct a supervised fine-tuning dataset.
Our dynamic decoding framework adaptively selects a strategy guided by an PU threshold. For low PU, it employs an aggressive strategy by selecting the label from the top-ranked vocabulary token; for high PU cases, it adopts a conservative strategy, sampling from the two tokens with the highest logits. To compute the PU, we first apply a ReLU activation to the token-level logits to derive their evidence. This evidence is then mapped to three canonical labels (HAWKISH, DOVISH, and NEUTRAL) based on a predefined token-to-label mapping. We construct a candidate evidence set by combining the aggregated label logits scores with the individual logits scores of all unmapped tokens. The top-$K$ logits scores from this set are then selected, and the PU is computed over them to capture ambiguity and low evidence. 

We utilized the open-source ModelScope Swift~\cite{zhao2025swift} framework for all model training and inference. All base models were fine-tuned using the LoRA method. During inference, we applied greedy decoding and incorporated FlashAttention for efficiency. For dynamic uncertainty decoding, hyperparameters were selected based on validation set performance. The search space was defined as follows: \textbf{Top-K:} \{3,\ 10,\ 15,\ 20,\ 25,\ 30\}; \textbf{Threshold percentiles:} \{1,\ 0.95,\ 0.9,\ 0.85,\ 0.8,\ 0.75,\ 0.7\}; \textbf{Sampling temperature:} \{0.1,\ 0.2,\ 0.3,\ 0.4,\ 0.5,\ 1.0,\ 1.5,\ 2.0\}. All stochastic components were controlled with a fixed random seed 42 to ensure reproducibility. All experiments were conducted on four NVIDIA A800 80GB GPUs. 
Full implementation details are provided in the appendix.

\subsection{Benchmark and Evaluation Metrics}
We train and evaluate our method on the  trillion dollar Words ~\cite{shah2023trillion} FOMC dataset, which contains three distinct categories of Federal Reserve communications: meeting minutes, press conference transcripts, and speeches. The dataset spans from January 1996 to October 2022, covering multiple economic cycles, including the dot-com bubble, the 2008 financial crisis, and the COVID-19 pandemic period. 
We report two metrics for evaluations and comparisons: Macro-F1 and Weighted-F1. Weighted-F1 is the original metric established for the FOMC dataset, ensuring consistency with previous research and enabling fair comparison with existing baselines. 

\subsection{Main Results}
\subsubsection{Performance Analysis} We compare our method with several widely used language models like GPT~\cite{achiam2023gpt}, Phi~\cite{abdin2024phi}, GLM~\cite{glm2024chatglm}, Qwen~\cite{yang2025qwen3}, Deepseek~\cite{guo2025deepseek}, and Gemini~\cite{comanici2025gemini}. The comparison also includes previous approaches like HD-Dissent~\cite{peskoff2023gpt}, AICBC~\cite{Fanta09042024}, and FinBERT~\cite{liu2021finbert}. For open-sourced language models, we report both zero-shot and fine-tuned results respectively. The results presented in Table \ref{table:main} demonstrate the effectiveness of our proposed approach across all three categories of Federal Reserve communications. 
Our proposed method achieves substantial improvements over all baselines, demonstrating the effectiveness of our approach for financial sentiment analysis. Specifically, our method achieves 0.7327 Macro-F1 and 0.7426 Weighted-F1 on the combined dataset, representing significant improvements of 6.6\% and 6.2\% respectively over the strongest baseline (0.6662 and 0.6802). These results highlight the effectiveness of our methods.

The performance gains are particularly pronounced for meeting minutes and speeches. On meeting minutes, our method achieves 0.7449 Macro-F1 and 0.7394 Weighted-F1, substantially outperforming the best baseline by 7.4\% and 6.5\% respectively. For speeches, we also observe improvements, with our method reaching 0.7291 Macro-F1 and 0.7718 Weighted-F1, representing gains of 6.7\% and 4.7\% over the best baseline.
However, in the press conference transcripts, our methods show reductions of 1.3\% and 2\% compared to GPT-4.1. The reason may be that press conferences involve real-time interactions where the context from prior questions and answers within the same session is critical for accurate sentiment classification. Our current approach may not fully capture these dynamic dependencies, which larger models like GPT-4.1 handle more effectively due to their enhanced contextual understanding.

\subsubsection{Fine-tuning vs. Zero-shot} An important observation from our results is the varying effectiveness of fine-tuning across different model architectures. While fine-tuning generally improves performance for most models, the magnitude of improvement varies significantly. For instance, GLM-4-9B shows substantial gains from fine-tuning (from 0.3152 to 0.6390 Macro-F1 on combined data), while larger models like Phi-4 show more modest improvements. This suggests that fine-tuning is particularly beneficial for models that may lack sufficient pre-training on financial domain data.
Notably, zero-shot models like GPT-4.1 achieve competitive performance without fine-tuning, indicating strong out-of-the-box capabilities for financial text understanding. However, our approach still outperforms these strong baselines.
\subsection{Ablation Analysis}
To better understand the contributions of various components in our proposed approach, we performed ablation studies by removing individual components step by step and evaluating their impact on performance in all categories. The results of ablation experiments are presented in Table \ref{table:abl}.

\begin{table}[ht]
\centering
\small
\begin{tabular}{lcc}
\toprule
  \multirow{2.5}{*}{\textbf{Model}} & \multicolumn{2}{c}{\textbf{All Categories}} \\ 
  \cmidrule{2-3}
  & \textbf{Macro F1} & \textbf{Weighted F1}\\
  \midrule
  Ours& 0.7327& 0.7426\\
  \quad w/o PU &0.7291 &0.7378\\
  \quad w/o Transmission Path& 0.6538&0.6699\\
  \quad w/o Entity Relationships& 0.6397 & 0.6551\\
  Original Qwen3 & 0.6360 &0.6534\\
  \bottomrule
  
\end{tabular}
\caption{Ablation study results of our methods.}
\label{table:abl}
\end{table}

Among the results, \textbf{w/o PU} refers to using greedy decoding for all samples, \textbf{w/o Transmission Path} indicates the removal of monetary policy transmission path information, \textbf{w/o Entity Relationships} means the model receives only the policy stance label guidelines, \textbf{original Qwen3} represents the baseline with all proposed modules removed. The transmission path shows the most substantial impact on model performance. Its removal results in the largest performance degradation. This significant drop demonstrates that explicit transmission paths are essential for complex financial sentiment analysis. 

The transmission path likely enables the model to analyze the sentiment by breaking down the decision process into interpretable steps, leading to more accurate and consistent predictions. The entity relationships also demonstrate substantial contribution to the model. It not only serves as the foundation for the transmission path but also brings performance improvement to the model. The uncertainty component shows the most modest improvement, however, the uncertainty quantification still provides valuable contributions by helping the model better calibrate its predictions and handle ambiguous cases more effectively.

\subsection{Uncertainty Analysis}
To validate the effectiveness of our uncertainty quantification algorithm, we conducted an analysis to examine the relationship between PU and prediction accuracy. We hypothesize that correctly predicted samples should exhibit lower PU compared to incorrectly predicted ones, which would demonstrate the reliability of our uncertainty estimation. We evaluated our uncertainty quantification approach in various $K$ values on the FOMC dataset. For each prediction, we calculated the PU based on the model outputs,  then classified predictions into correct and incorrect groups, and computed the mean PU for each group.

To rigorously test our hypothesis, we employed three complementary statistical methods:
\begin{itemize}
\item \textbf{T-test}: To compare mean PU between correct and incorrect predictions
\item \textbf{Mann-Whitney U test}: A non-parametric alternative to validate results without assuming normal distribution
\item \textbf{Logistic regression}: To quantify the predictive power of PU for classification accuracy
\end{itemize}

\begin{figure}[ht]
\centering
\includegraphics[width=0.98\linewidth]{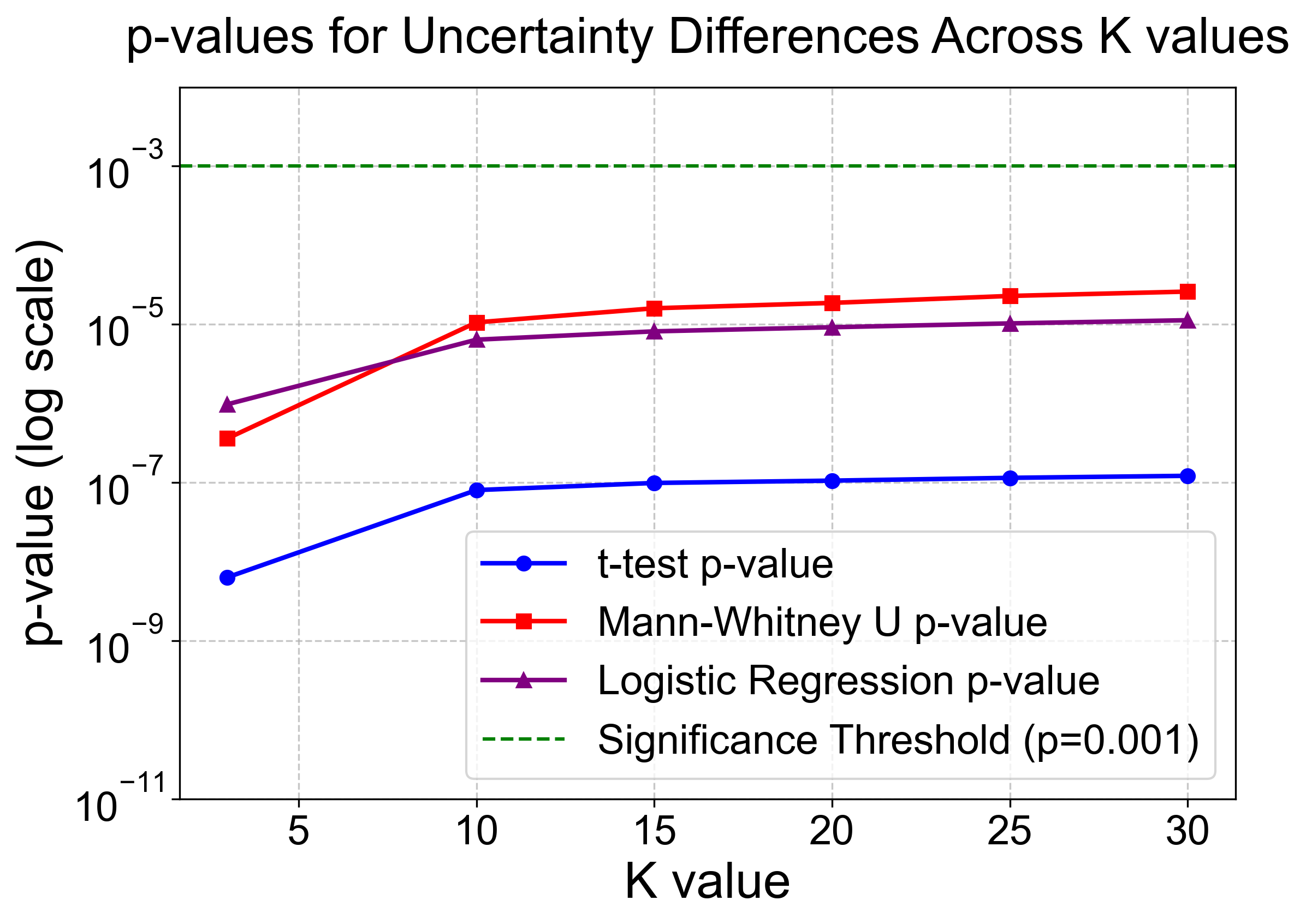}
\caption{P-values of T-test, Mann-Whitney U test and logistic regression for different $K$ values on FOMC dataset.}
\label{fig:uncertainty_pvalues}
\end{figure}

Test results are shown in Figure \ref{fig:uncertainty_pvalues}. Both statistical tests yield consistently significant results across all k-values. The blue line shows p-values of T-test ranging from $6 \times 10^{-9}$ to $1.2\times10^{-7}$, all substantially below the significance threshold (green dashed line at p=0.001). The red line demonstrates p-values of the Mann-Whitney U test from $4\times10^{-7}$ to $3\times10^{-5}$, providing non-parametric confirmation of significant differences and ensuring robustness regardless of data distribution assumptions. The purple line represents p-values of logistic regression from $1 \times 10^{-6}$ to $1.2\times10^{-5}$, indicating that PU serves as a significant predictor of correctness.

The above results confirm a significant positive correlation between model PU and model error rate.
This finding provides crucial guidance for practical applications: when the model's PU is high, the reliability of its predictions correspondingly decreases.
In real-world financial sentiment analysis scenarios, the most prudent strategy is to actively seek human expert intervention or choose to abstain from providing an answer when the model detects that it cannot make accurate judgments (i.e., when the PU exceeds threshold). This strategy can effectively prevent potential losses caused by incorrect predictions, particularly in high-risk financial decision-making environments.

However, to ensure fair comparison with existing baseline methods, we did not provide models with the option to ``refuse to answer" in our main experiments. Instead, all models were required to provide explicit predictions for every sample.
To further validate the effectiveness of uncertainty quantification in practice, we designed additional experiments to demonstrate the accuracy differences when model predictions have PU above or below thresholds. We list the accuracy of the low PU group and the high PU group, respectively, in Table \ref{table:cert}.

\begin{table}[ht]
\centering
\small
\setlength{\tabcolsep}{6mm}
\begin{tabular}{lcc}
\toprule
  \multirow{2.5}{*}{\textbf{PU}} & \multicolumn{2}{c}{\textbf{All Categories}} \\ 
  \cmidrule{2-3}
  & \textbf{Macro F1} & \textbf{Weighted F1}\\
  \midrule
  Low  & 0.7791 & 0.7822\\
  High & 0.2473 &0.4372\\
  \bottomrule
  
\end{tabular}
\caption{Performance of predictions with low vs. high PU on the FOMC dataset.}
\label{table:cert}
\end{table}

These results indicate that predictions with low PU are highly reliable, achieving strong performance across both Macro F1 and Weighted F1 metrics. In contrast, the high PU group exhibits significantly poorer performance, indicating a higher likelihood of errors. This large performance gap highlights that high PU predictions negatively impact overall model accuracy, effectively undermining the robust results of the low PU group. These findings further reinforce the positive correlation between PU and prediction error rate, confirming PU as a reliable signal of prediction reliability.


\section{Case Study}
\begin{table}[htbp]
\small
\centering
\begin{tabularx}{\columnwidth}{>{\raggedright\arraybackslash}X l l} 
\toprule
\textbf{Sentence Type} & \textbf{Prediction}  & \textbf{Strategy} \\
\midrule
Explicit + Contrastive & Correct  & \textbf{Aggressive} \\
\addlinespace 
Contextual Confusion & Incorrect  & \textbf{Aggressive} \\
\addlinespace 
Implicit Statement & Incorrect  & \textbf{Conservative} \\
\bottomrule
\end{tabularx}
\caption{Case study of different sentence types.}
\label{tab:case_study_accuracy}
\end{table}
We conduct a case study to analyze model behavior under different sentence types. Results are shown in Table~\ref{tab:case_study_accuracy}. The model performs well when processing explicit signals, especially those with contrastive shifts (e.g., but, however), where semantic redirection is clear. It also correctly captures transitional tones when contextual cues are clear.
However, the model struggles in two cases. First, under contextual confusion, even when surface keywords are present, the sentence may reflect third-party views or rely on external background knowledge. Without such knowledge, the model misinterprets the sentence and makes overly aggressive decisions. Second, under implicit statements, where no clear keywords or contextual anchors exist, the model fails to resolve the underlying intent through economic reasoning. In this case, it adopts a conservative strategy but still produces incorrect predictions.
Table~\ref{tab:case_study_accuracy} categorizes these cases by sentence type, prediction correctness, and decision strategy.
The original sentences and detailed analysis for case study are provided in the detailed case study section of the appendix.


\section{Conclusion}
We present a domain-specific reasoning framework grounded in the monetary policy transmission mechanism to emulate expert analysis of policy stances. 
We propose a dynamic uncertainty decoding module that effectively identifies cases where the model lacks sufficient knowledge or exhibits contextual confusion, rendering it unable to provide viable predictions. Under high perceptual uncertainty, our approach enables the model to identify unreliable outputs, improving reliability under realistic decision-making conditions.
We validate our approach through a comprehensive hyperparameter search and demonstrate that the framework achieves state-of-the-art performance on the policy stance analysis task. Beyond improved robustness, our framework also supports human analysts by highlighting predictions that are well-grounded versus those requiring caution.
Our work offers a new direction for evaluating the reliability, transparency, and economic interpretability of LLM predictions in policy stance analysis.

\section{Limitation}
While incorporating domain knowledge and dynamic decoding enhances performance and reliability, our approach still relies on hand-crafted templates and has runtime and strategy constraints. The main limitation is the limited availability of fine-grained data from central banks such as the BoE and ECB. Additionally, the framework needs improvement to better handle context ambiguity and implicit statements.
We leave these challenges for future work.

\section{Acknowledgments}
This work is supported by the National Natural Science Foundation of China (No. 62406267) and the Guangzhou Municipal Science and Technology Project (No. 2025A04J4070). Zhang acknowledges financial support from the National Science Foundation of China (No. 72373097).
\bibliography{aaai2026}

\clearpage
\appendix
\section{Appendix}
\subsection{Data Augmentation \& Implementation}
\subsubsection{Policy stance Definition}
We provide additional notes on our definitions and explanations of policy stance labels as follows:

\begin{table}[htbp]
\centering
\small 
\begin{tabularx}{\columnwidth}{>{\bfseries\RaggedRight}p{0.2\columnwidth}>{\RaggedRight}X}
\toprule
\multicolumn{1}{c}{\textbf{Label}} & \multicolumn{1}{c}{\textbf{Definition}} \\
\midrule

\textbf{DOVISH} 
& \compactcell{
     Dovish tends towards an accommodative monetary policy, supporting lower interest rates, quantitative easing, and other stimulus measures to promote economic growth and employment. \\
} \\
\addlinespace[1mm]

\textbf{NEUTRAL} 
& \compactcell{
    There is no clear inclination in the stance of monetary policy. \\
} \\
\addlinespace[1mm]

\textbf{HAWKISH} 
& \compactcell{
    Hawkish tends to adopt a contractionary monetary policy, placing more emphasis on controlling inflation, even at the cost of slowing economic growth. 
} \\
\bottomrule
\end{tabularx}
\caption{Definition of monetary policy stance labels.} 
\label{tab:monetary_policy}
\end{table}
In the definition of a dovish stance, two levels are included. First, the text expresses an explicit signal that the Federal Reserve favors an accommodative monetary policy. Second, the text content implies that the Fed tends to adopt an accommodative monetary policy (Neutral-Dovish). It can be derived from corresponding economic phenomena that an accommodative monetary policy might be adopted, or it directly expresses related concerns, but still requires further data support or other statements for cross-validation. And in the definition of a Hawkish stance, the text expresses an explicit signal that the Fed tends to adopt a contractionary monetary policy. Besides, the text content implies that the Fed tends to adopt a contractionary monetary policy (Neutral-Hawkish). It can be derived from corresponding economic phenomena that a contractionary monetary policy might be adopted, or it directly expresses related concerns, but still requires further data support or other statements for cross-validation. Neutral stance indicates that the text is a neutral description, with no clear loose or tight tendency.\\

\subsubsection{Financial Entity Relations}
Our six core relationships are shown in Table~\ref{tab:Core Relations}.
\begin{table}[htbp]
\centering

\small 
\begin{tabularx}{\columnwidth}{>{\bfseries\RaggedRight}p{0.2\columnwidth}>{\RaggedRight}X}
\toprule
\multicolumn{1}{c}{\textbf{Core relations}} & \multicolumn{1}{c}{\textbf{Definition}} \\
\midrule

\textbf{CAUSE} 
& \compactcell{
    Entity A leads to or results in a change in Entity B (Causal) \\
} \\
\addlinespace[1mm]

\textbf{COND} 
& \compactcell{
    Entity A establishes a necessary precondition for Entity B (Conditional) \\
} \\
\addlinespace[1mm]

\textbf{EVID} 
& \compactcell{
    Entity A provides evidence for or supports a conclusion about Entity B (Evidential) \\
} \\
\addlinespace[1mm]

\textbf{PURP} 
& \compactcell{
    The goal or purpose of Entity A is to achieve Entity B (Purpose) \\
} \\
\addlinespace[1mm]

\textbf{ACT} 
& \compactcell{
    Entity A (the agent) performs the action B (the verb/event) (Action) \\
} \\
\addlinespace[1mm]

\textbf{COMP} 
& \compactcell{
    Entity A is being compared with Entity B (Comparative) 
} \\
\bottomrule
\end{tabularx}
\caption{Definition of core relations.} 
\label{tab:Core Relations}
\end{table}

\subsubsection{Implementation}
We implement the data augmentation process in three steps. First, we use an entity-relation extraction template to prompt the LLM to generate a list of economically relevant reasoning tasks. These tasks are structured as subtasks associated with identified entities.

Second, each subtask is passed to the LLM responsible for transmission path reasoning. Based on the original text and the decomposed subtasks, the LLM generates candidate economic inferences and policy suggestions using a transmission path template. This results in a set of intermediate reasoning paths and policy implications.

Third, to produce the final explanation and final reasoning path, we use the shared policy stance analysis template. On the training set, we first use the previously generated entity relations and transmission paths to prompt the LLM. If the predicted labels differ from the ground truth, we invoke the LLM again to align the label, followed by human correction of both the explanation and reasoning path by two annotators. For predictions that match the ground truth, we still perform manual verification to ensure consistency and clarity. We employed data augmentation using Deepseek-R1-0528. The structured policy stance analysis prompt templates is provided with the code.
\subsubsection{Entity Extraction Template Design}
We design a task planner prompt to instruct the LLM to identify reasoning units from financial policy text and generate corresponding sub-tasks. The planner performs two main functions: extracting key financial entities and generating reasoning relations using economic logic. 
The simplified form of the prompt template is shown in listing~\ref{lst:planner-prompt}.
\begin{listing}[htb!]
\begin{lstlisting}
########################################
system prompt: You are a professional financial analyst skilled in identifying entities and their relationships in financial texts.
########################################
**ROLE**: You are a professional financial analyst specialized in FOMC statement analysis.
Your core mission is twofold:1. Entity Recognition: Identify key financial entities and their sources from the text. 2. Relation Extraction: Map entity interactions using 6 core relations + 2 composite patterns.

Core Relations & Composite Patterns:
(This part is align with table Definition of core relations)

Predefined category template:
{category_template}

Key Annotation Principles & Rules:
Rule A: Entity Grouping & Classification
Rule B: Logical Chains & Decomposition
Rule C: Adversarial Structure (BUT)
Rule D: Source Attribution

Example Comprehensive Relation:
(((productivity growth slowed + employment picked up) COND (reductions in slack)) CAUSE (higher unit labor costs)) CAUSE pressures on prices

Example decomposition into atomic relations:
(A + B) COND C
C CAUSE D
D CAUSE E

Output fomrmat:
......
\end{lstlisting}
\caption{Core Logic of the Planner Prompt for Entity-Relation Extraction}
\label{lst:planner-prompt}
\end{listing}

\subsubsection{Transmission Path Reasoning Template Design}
The transmission path reasoning prompt is designed to emulate the analytical process of human experts. It takes structured entity relations and text snippets as input and infers possible monetary policy transmission paths. The template consists of three key components: relevance assessment, structured reasoning via economic transmission channels, and policy advice generation.
The simplified form of the prompt template is shown in listing~\ref{lst:Executor-prompt}.
\begin{listing}[htb!]
\begin{lstlisting}
########################################
system prompt: You are a monetary policy transmission path analyst.
########################################
**ROLE**: You are a monetary policy transmission path analyst. Your task is to infer potential monetary policy transmission paths based on provided entity relations and text snippets.

# Monetary Policy Transmission Path CORE DEFINITION
## Task Requirements
Economic Relevance Gatekeeping: (This part is to filter non-substantive signals)

Definition of X, Y, Z, M

Transmission Paths Template:
### Format A: X (Phenomenon) -> Z (Changes in economic/market indicators -> Market participants revise policy expectations) -> M (Policy recommendation)
Use this when the market/economic indicators respond first, and the expectation of policy reaction arises later as a result of these observable shifts.
### Format B: X (Phenomenon) -> Z (Immediate revision of policy expectations -> Changes in market/economic indicators) -> M (Policy recommendation)
Use this when the policy expectation shifts immediately, and the market indicators respond later, often through asset repricing or economic behavior adjustments.

Guidence for Transmission Path Building:
......

Few-shot Example:
......

Output Fomrmat:
......
\end{lstlisting}
\caption{Core Logic of the Executor Prompt for Monetary Policy Transmission Path Reasoning}
\label{lst:Executor-prompt}
\end{listing}

\subsubsection{Policy Stance Analysis Template Design}
The policy stance analysis prompt is designed to simulate human-style monetary policy stance annotation by integrating multiple information sources. It determines the overall stance (DOVISH, HAWKISH, or NEUTRAL) based on the original text, structured financial entity relations, metadata, and inferred policy transmission paths. 
The simplified form of the prompt template is shown in listing~\ref{lst:Finalizer-prompt}.
\begin{listing}[htb!]
\begin{lstlisting}
########################################
system prompt: You are a professional policy stance analyst. Your task is to determine the overall policy stance conveyed by the provided original text and respond strictly in the specified JSON format.
########################################
**ROLE**: You are a professional financial sentiment analyst. Your task is to determine the overall financial sentiment tendency expressed in the original sentence/paragraph based on the provided original text, financial entity relationships, metadata (time, source of text), and monetary policy transmission paths.

# Input information: original text, financial entity relations, metadata, monetary policy transmission path, human annotation guideline.

# Policy Stance Label Definition:
......

# Output Format:
......
\end{lstlisting}
\caption{Core Logic of the Policy Stance Analysis Prompt}
\label{lst:Finalizer-prompt}
\end{listing}

\subsection{Detailed Experiment Setup}
The detailed scripts, codes and system environment are included in the supplementary code package.
\subsubsection{Supervised Fine-tuning Experiment}
We conducted LoRA-based supervised fine-tuning using the MS-Swift framework. The key hyperparameters are summarized as follows:
We set the LoRA rank to 8 and the scaling factor (lora-alpha) to 32. The model was trained for 5 epochs. We used a per-device batch size of 1 with gradient accumulation over 2 steps on 4 GPUs, resulting in an effective batch size of 8 for parameter updates. We used the bfloat16 precision and enabled gradient checkpointing to reduce memory consumption. The optimizer used a learning rate of 1e-4. All linear layers were included as target modules for LoRA adaptation. During training, model checkpoints were saved every 60 steps, and validation was also performed every 60 steps. We retained up to 12 checkpoints. Data loading was parallelized using 4 worker threads. We disabled sequence packing and fixed the random seed to 42 to ensure reproducibility. All training was conducted with DeepSpeed ZeRO-3 optimization.

\subsubsection{Dynamic Uncertainty Decoding Workflow}
To accelerate hyperparameter tuning for dynamic uncertainty decoding, we designed a two-step inference workflow. In the first step, we identified candidate token sets at the policy stance prediction position for each input based on predefined patterns. These token sets were extracted and saved, with a one-to-one mapping maintained between each input and its corresponding candidate list. In the second step, we reused pre-computed model logits and performed efficient hyperparameter search over the saved candidate positions without re-running the full decoding process. To ensure consistency and reproducibility of reported results, all performance metrics are based on outputs from the hyperparameter search workflow.

\subsection{Ablation Analysis}
\subsubsection{Data Augmentation Ablation}
The ablation study is conducted with Gemini-2.5-Pro under an end-to-end evaluation setup.

\begin{table}[ht]
\centering
\begin{tabular}{l c}
\toprule
\textbf{Setting} & \textbf{Weighted-F1} \\
\midrule
Full augmentation prompt & \textbf{0.6695} \\
w/o rules                & 0.5073 \\
w/o examples             & 0.5024 \\
w/o both                 & 0.4950 \\
\bottomrule
\end{tabular}
\caption{Ablation results for the augmentation prompt. Truncating or simplifying the template consistently reduces accuracy.}
\label{tab:augmentation_ablation}
\end{table}

The ablation results show that both rules and examples play essential roles in the augmentation prompt. Removing either component leads to a substantial drop in Weighted-F1, and removing both results in the lowest performance. The full prompt provides the strongest guidance and produces the most reliable augmented data, confirming that the structured template is necessary for stable model behavior.
\subsubsection{Policy Stance Label Definition Strategy}
We reviewed prior work on policy stance prediction with LLMs and found that label definitions generally follow two approaches: using semantically meaningful textual labels (e.g., dovish, hawkish, neutral), and mapping stance categories to numeric labels (e.g., 1, 0, –1). To examine how label definition affects model performance, we conducted an ablation study by fine-tuning models on datasets with different label formats. As shown in Table~\ref{tab:label_def_ablation}, using original text-based labels yields a significantly higher weighted F1 score 0.7378 compared to using numeric labels (0.7114).
This performance gap may be attributed to the semantic alignment between the original textual labels and the prediction task. Text-based labels provide clearer learning signals and reduce the need for the model to internally remap abstract numbers to policy semantics, thereby lowering the learning burden.
\begin{table}[ht]
\small
\centering
\begin{tabular}{lcc}
\toprule
\textbf{Label Type} & \textbf{Macro F1} & \textbf{Weighted F1} \\
\midrule
Numeric Label & 0.7024 & 0.7114 \\
Text-based Label & \textbf{0.7291} & \textbf{0.7378} \\
\bottomrule
\end{tabular}
\caption{Performance comparison: numeric vs. text-based policy stance label definitions.}
\label{tab:label_def_ablation}
\end{table}

\subsubsection{Token Vocabulary}
In the dynamic uncertainty decoding experiment, we observed that the evidence for a given label can be fragmented across multiple candidate token IDs due to the behavior of the model's tokenizer. Specifically, we found that for the same label semantics, the Qwen-3 produces fragmented token outputs similar to the English-centric Phi-4 model. The candidate token vocabularies and their associated logit scores are visualized in Table~\ref{tab:token_vocab_logit_compare}.

\begin{table}[ht]
\centering
\begin{tabular}{c cc cc}
\toprule
\multirow{2}{*}{\textbf{Rank}} & \multicolumn{2}{c}{\textbf{Qwen3-14B}} & \multicolumn{2}{c}{\textbf{Phi-4}} \\
\cmidrule(lr){2-3} \cmidrule(l){4-5}
& \textbf{Token} & \textbf{Logit} & \textbf{Token} & \textbf{Logit} \\
\midrule
1  & H        & 24.2500 & H        & 24.3750 \\
2  & NE       & 22.2500 & NE       & 24.0000 \\
3  & DO       & 21.1250 & DO       & 21.6250 \\
4  & hawk     & 17.5000 & hawk     & 16.8750 \\
5  & HA       & 16.2500 & NA       & 16.8750 \\
6  & NA       & 15.8750 & Neutral  & 16.1250 \\
7  &  Hawk    & 15.6250 & HO       & 15.7500 \\
8  &  hawk    & 15.0000 & Do       & 15.6250 \\
9  & HO       & 14.9375 & HA       & 15.6250 \\
10 & HE       & 14.8750 & E        & 15.3750 \\
11 & HAL      & 14.6250 & HE       & 15.3750 \\
12 & HI       & 14.5625 & UNKNOWN  & 15.3125 \\
13 & Ne       & 14.3125 & UN       & 14.8125 \\
14 & Neutral  & 14.1875 & ne       & 14.7500 \\
15 & DA       & 14.0625 & BE       & 14.4375 \\
16 & hawks    & 14.0000 & NO       & 14.3125 \\
17 & HW       & 13.9375 & N        & 14.3125 \\
18 & h        & 13.6875 & do       & 14.3125 \\
19 & N        & 13.6250 & Ne       & 14.1875 \\
20 & D        & 13.4375 & HAV      & 14.1875 \\
\bottomrule
\end{tabular}
\caption{Prediction Token Vocabulary (Top-20 Tokens by Logit Score).}
\label{tab:token_vocab_logit_compare}
\end{table}

We conducted a comparative analysis between our method and a baseline approach that directly applies top-K selection over candidate tokens without grouping. Experimental statistics show that, when K is small (K = 3, Mann-Whitney test, p=0.0628), the baseline method exhibits weaker uncertainty significance. As K increases, its uncertainty scores become more strongly correlated with prediction error rates. Based on this observation, we further tuned K on the validation set and reported test results using the best-performing configuration, as shown in Table~\ref{tab:uncertainty_cluster_compare}.

\begin{table}[ht]
\small
\setlength{\tabcolsep}{4pt}  
\centering
\begin{tabular}{lcccccc}
\toprule
\textbf{Strategy} & \textbf{K} & \textbf{Thres.} & \textbf{Temp.} & \textbf{Macro F1} & \textbf{W. F1} \\
\midrule
Non-clustered        & 20 & 0.9 & 0.4 & 0.7321 & 0.7404 \\
Vocab-clustered      & 10 & 0.8 & 0.4 & \textbf{0.7327} & \textbf{0.7426} \\
\bottomrule
\end{tabular}
\caption{Comparison of uncertainty-based strategies with vs. without label clustering.}
\label{tab:uncertainty_cluster_compare}
\end{table}

While the baseline method benefits from larger K values, its performance remains inferior to our proposed approach. We attribute this to the tokenizer-induced token-level dispersion, which causes evidence for the true label to be fragmented across multiple token IDs. This dispersion weakens the model's preference signal for the correct label and leads to less reliable uncertainty estimation. 

\subsubsection{Conservative and Aggressive Strategy Design}
Based on our proposed PU metric, we designed a set of inference strategies to evaluate model behavior under uncertainty.
For aggressive strategies, we implemented the candidate-greedy (greedy-c) strategy reported in the main text. We also designed an additional aggregate-K-greedy (greedy-k) strategy, which selects the top-1 token from the grouped candidate vocabulary based on label clustering.
For conservative strategies, we did not include a ``refuse to answer" option to ensure a fair comparison with existing baselines. Instead, we default to predicting a neutral stance in high-PU scenarios to simulate abstention.
To test whether our method enhances the model's ability to utilize label-specific evidence during prediction, we further implemented two sampling-based variants:
\begin{itemize}
    \item \textbf{candidate sampling}, which samples from the top-2 tokens in the original candidate vocabulary;
    \item \textbf{cluster sampling}, which samples from the top-2 tokens within each grouped label cluster.
\end{itemize}
We conducted a series of experiments using the above strategies. Statistical results show that both the greedy-k and greedy-c strategies exhibit a significant positive correlation between PU scores and prediction error rates across the range of K values. We further applied each strategy to search for its optimal hyperparameter on the validation set and evaluated the best configuration on the test set. The results are summarized in Table~\ref{tab:strategy_f1}.
\begin{table}[t]
\centering
\small
\setlength{\tabcolsep}{2.8pt}  
\begin{tabular}{llccc|cc}
\toprule
\multirow{2}{*}{Aggressive} & \multirow{2}{*}{Conserv.} & \multirow{2}{*}{K} & \multirow{2}{*}{Thres.} & \multirow{2}{*}{Temp.} & \multicolumn{2}{c}{F1} \\
\cmidrule(lr){6-7}
& & & & & Macro & Weighted \\
\midrule
\multirow{3}{*}{greedy-k} 
& cand. & 3 & 0.90 & 0.4  & 0.7287 & 0.7363 \\
& clst. & 3 & 0.75 & 2.0  & 0.7273 & 0.7366 \\
& NEU.  & 3 & 0.75 & --   & 0.7269 & 0.7364 \\
\midrule
\multirow{3}{*}{greedy-c} 
& cand. & 10 & 0.80 & 0.4  & \textbf{0.7327} & \textbf{0.7426} \\
& clst. & 3  & 0.75 & 2.0  & 0.7273 & 0.7366 \\
& NEU.  & 3  & 0.75 & --   & 0.7269 & 0.7364 \\
\bottomrule
\end{tabular}
\caption{Test set performance of strategy combinations using best validation-tuned hyperparameters. “--” means no temperature used in NEUTRAL.}
\label{tab:strategy_f1}
\end{table}
As shown in Table~\ref{tab:strategy_f1}, the best-performing hyperparameters for both the greedy-k strategy and the conservative strategies (cluster and neutral) consistently use K = 3. However, these configurations fail to outperform the baseline weighted F1 score of 0.7378 on the test set. This may indicate that the cluster and neutral strategies suffer from overfitting to the validation set, resulting in weaker generalization and reduced robustness.
In contrast, the greedy-c strategy combined with candidate sampling achieves the highest test performance, with a weighted F1 of 0.7426.
\subsubsection{Transferability Evaluation}
We further conducted transferability experiments, as shown in Table~\ref{tab:transferability}. The results demonstrate that our method retains a consistent performance advantage across models, indicating its transferability.

\begin{table}[ht]
\small
\centering
\begin{tabular}{lcc}
\toprule
\textbf{Model} & \textbf{Macro F1} & \textbf{Weighted F1} \\
\midrule
Phi-4 (Ours)     & \textbf{0.7147} & \textbf{0.7241} \\
Phi-4 (Baseline) & 0.7139          & 0.7225          \\
\midrule
Qwen-3 (Ours)     & \textbf{0.7327} & \textbf{0.7426} \\
Qwen-3 (Baseline) & 0.7291          & 0.7378          \\
\bottomrule
\end{tabular}
\caption{Transferability test using best hyperparameters searched on Qwen-3 (K=10, Threshold=0.8, Temperature=0.4).}
\label{tab:transferability}
\end{table}

This analysis suggests that the our proposed method is more effective in leveraging PU scores to guide predictions and improve reliability. Subsequent sensitivity analysis of hyperparameter settings further support the robustness of this approach.

\subsection{Hyperparameters Sensitivity Analysis}
\begin{figure}[ht!]
    \begin{subfigure}{\linewidth}
        \centering
        \includegraphics[width=0.9\textwidth]{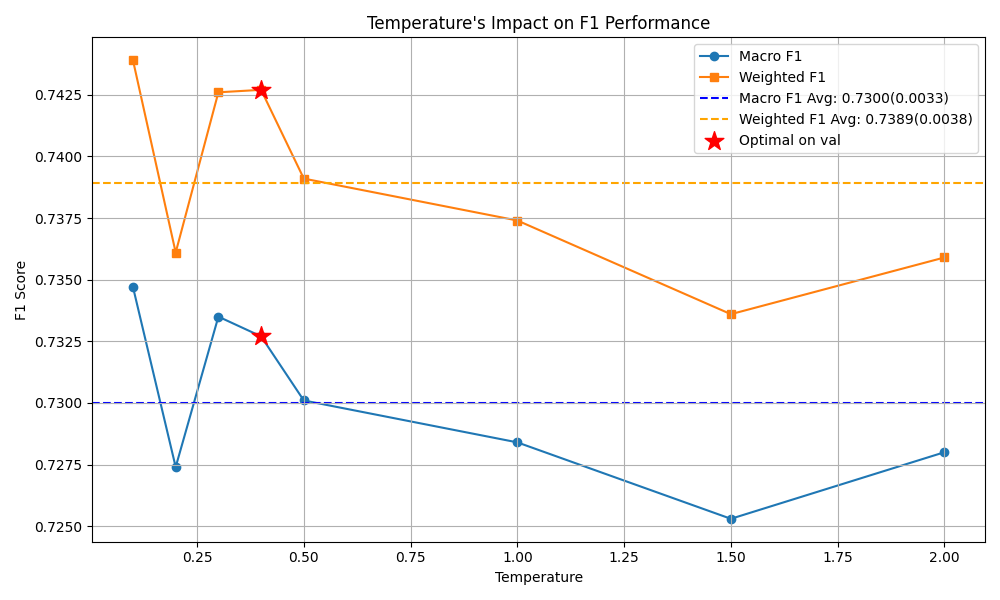}
        \caption{Temperature's Impact on F1 Performance.}
        \label{subfig:temperature}
    \end{subfigure}

    \begin{subfigure}{\linewidth}
        \centering
        \includegraphics[width=0.9\textwidth]{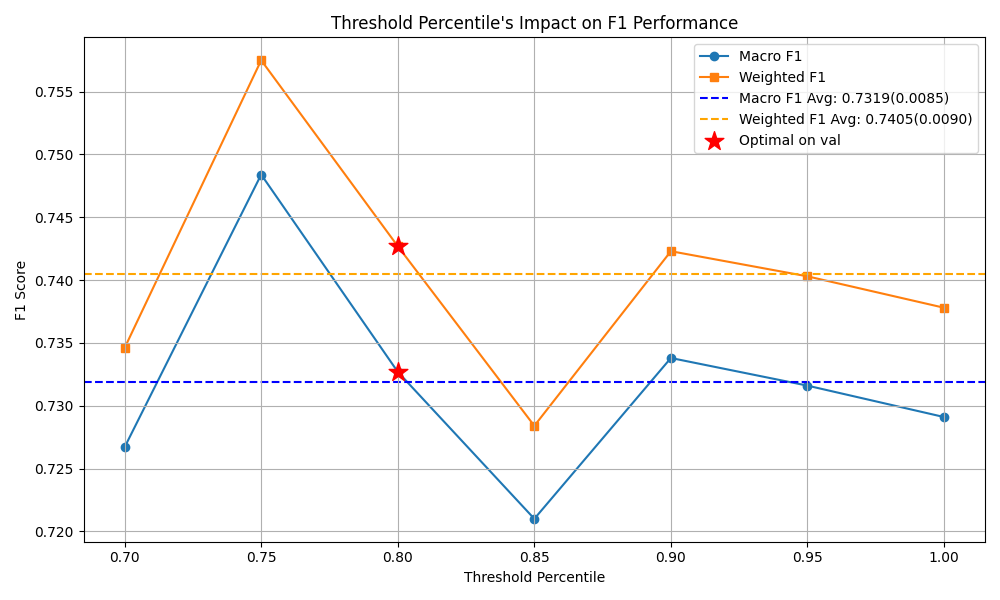}
        \caption{Threshold Percentile's Impact on F1 Performance.}
        \label{subfig:threshold}
    \end{subfigure}

    \begin{subfigure}{\linewidth}
        \centering
        \includegraphics[width=0.9\textwidth]{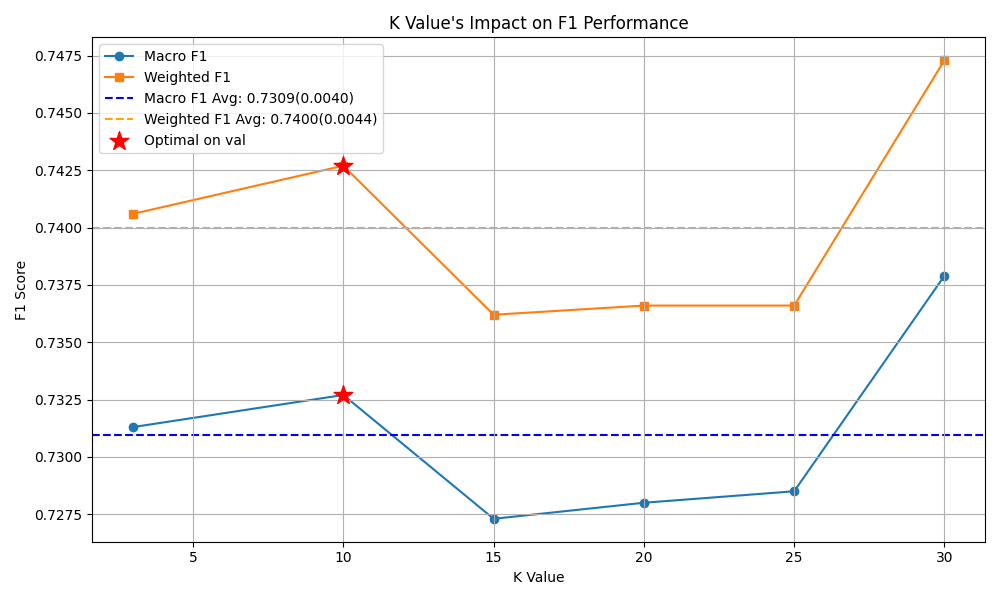}
        \caption{K value's Impact on F1 Performance.}
        \label{subfig:k_value}
    \end{subfigure}
    
    \caption{Comparison of F1 Performance across Temperature, Threshold Percentile and K value.}
    \label{fig:combined}
\end{figure}

This section analyzes the sensitivity of the model's performance to variations in three key hyperparameters: temperature, threshold percentile, and k-value. The performance metrics considered are the macro F1 score and the weighted F1 score. For each hyperparameter group, we report the average of the results and the standard deviation. 
\subsubsection{Temperature} As shown in Figure \ref{subfig:temperature}, the overall mean for macro F1 across all temperatures is 0.7300 with a standard deviation of 0.0033, and for weighted F1, the mean is 0.7389 with a standard deviation of 0.0038. The low standard deviations indicate stable performance across the temperature range, with the best results at lower temperatures.
\subsubsection{Threshold Percentile} Figure \ref{subfig:threshold} contains the threshold percentile’s impact on F1 performance. The overall mean for macro F1 across all thresholds is 0.7319 with a standard deviation of 0.0085, and for weighted F1, the mean is 0.7405 with a standard deviation of 0.0090. Most results are clustered around the mean, except for one maximum value (0.75) and one minimum value (0.85).
\subsubsection{K Value} We show the k value's impact on F1 performance in \ref{subfig:k_value}. The overall mean for macro F1 across all k-values is 0.7309 with a standard deviation of 0.0040, and for weighted F1, the mean is 0.7400 with a standard deviation of 0.0044. Notably, the highest performance is observed at a k-value of 30, indicating that relatively large k-values can be beneficial to the model.

In conclusion, our model is robust to variations in temperature and k-value, allowing flexibility in these settings. The threshold percentile is the most influential parameter, as it produces the largest performance swings and higher standard deviations. Additionally, although our optimal hyperparameters on validation set fail to obtain the strongest performance on test set, all of them outperform the average of the results.
\subsection{Case Study}
We provide supplementary notes for the case study section in the main text. In Table~\ref{tab:Case study with sentence classification and prediction analysis}, we present full sentences with their sentence types, GT labels, prediction labels, and correctness. We also include relevant analysis.

\begin{table*}[t] 
\small
\centering
\renewcommand{\arraystretch}{1.25} 
\setlength{\tabcolsep}{4pt} 
\begin{tabularx}{\textwidth}{ 
  >{\raggedright\arraybackslash}p{0.18\textwidth} 
  >{\centering\arraybackslash}p{0.10\textwidth} 
  >{\centering\arraybackslash}p{0.10\textwidth} 
  >{\centering\arraybackslash}p{0.10\textwidth} 
  >{\centering\arraybackslash}p{0.10\textwidth} 
  >{\raggedright\arraybackslash}X 
}
\toprule
\textbf{Sentence} & \textbf{Sentence Type} & \textbf{GT Label} & \textbf{Prediction Label} & \textbf{Correctness} & \textbf{Analysis} \\
\midrule
\textbf{Sentence 1:} I would say, in the area of commercial real estate, while valuations are high, we are seeing some tightening of lending standards and less debt growth associated with that rise in commercial real estate prices. 
& Explicit + Contrastive & Hawkish & Hawkish & Correct 
& We observe that although the sentence lacks explicit contrast-signaling words like ``but" or ``however", the statement ``we are seeing some..." logically implies a contrast with the preceding statement ``while valuations are high". Our model establishes a risk-averse stance that shapes the hawkish outcome through the causal emphasis in the BUT relationship. \\
\addlinespace[3pt]

\textbf{Sentence 2:} This is perhaps because the emphasis on price stability is taken by some as carrying a hint of restrictive policy and as an inclination to always be leaning against cyclical increases in demand.
& Contextual Confusion & Neutral & Hawkish & Incorrect 
& We note that the statement   ``This is perhaps because..." suggests an explanatory and descriptive tone, rather than a direct assertion by the Federal Reserve; it conveys a neutral policy stance. However, when interpreting the Fed's dialogue, our model focused solely on key evidence like ``emphasis on price stability," ``restrictive policy," and ``leaning against cyclical increases in demand," without considering the context. Our model concluded that a tightening monetary policy was warranted, indicating a hawkish stance. In this context, contextual confusion led to an incorrect judgment.  \\
\addlinespace[3pt]

\textbf{Sentence 3:} Yet, with a few exceptions, the available data show that productivity growth in other advanced countries has not increased to the extent seen in the United States.
& Implicit Statement & Dovish & Neutral & Incorrect 
& We find that this sentence argues against rushing to tighten policy in the US by emphasizing external economic weakness, indirectly supporting loose monetary policy. The phrases ``with a few exceptions" and ``not increased to the extent" highlight universal weakness, not positive exceptions. This sentence shows a clear dovish tilt. From this perspective, our model failed to recognize the implicit policy message: that the US comparison with other countries conveys a dovish stance. Our model thus interpreted it as neutral, leading to an incorrect judgment.  \\
\bottomrule
\end{tabularx}
\caption{Comprehensive case study with sentence classification and prediction analysis.}
\label{tab:Case study with sentence classification and prediction analysis}
\end{table*}

\end{document}